\documentclass[a4paper,10pt]{article}
\usepackage{graphicx} 
\usepackage{url}
\newcommand{\Roff}{R_{\mathrm{off}}}
\newcommand{\Ron}{R_{\mathrm{on}}}
\begin{document}
%
%
\title{Comparison of Ant-Inspired Gatherer Allocation Approaches using Memristor-Based Environmental Models}
%
%
\author{Ella Gale$^1$, Ben de Lacy Costello$^1$ \and Andrew Adamatzky$^1$\\
\\
1. Unconventional Computing Group,\\ 
University of the West of England, Bristol, UK,\\
ella.gale@uwe.ac.uk\\
\texttt{http://uncomp.uwe.ac.uk/index.html}}
%
%
%


\begin{abstract}        
Memristors are used to compare three gathering techniques in an already-mapped environment where resource locations are known. The All Site model, which apportions gatherers based on the modeled memristance of that path, proves to be good at increasing overall efficiency and decreasing time to fully deplete an environment, however it only works well when the resources are of similar quality. The Leaf Cutter method, based on Leaf Cutter Ant behaviour, assigns all gatherers first to the best resource, and once depleted, uses the All Site model to spread them out amongst the rest. The Leaf Cutter model is better at increasing resource influx in the short-term and vastly out-performs the All Site model in a more varied environments. It is demonstrated that memristor based abstractions of gatherer models provide potential methods for both the comparison and implementation of agent controls.
\paragraph{keywords:} memristor memristors networks gathering model multi-agent systems ants leaf-cutter Atta gatherer animal behaviour

\end{abstract}

\maketitle  

\section{Introduction}

\subsection{Memristors}

Memristors are an emerging technology with anticipated wide-spread applications in neuromorphic computing, artificial intelligence and green technology. The memristor is the 4th fundamental circuit element predicted to exist in 1971~\cite{Chua1971} and was only brought to wide-spread attention in 2008~\cite{Nature2008}. A memristor differs from the resistor by being able to store a state ie. it possesses a memory. Standard modern-day computers separate the processor and memory to different physical places, whereas the memristor can be used as both. As this is similar to how the brain works~\cite{L-B2009}, it is thought that artificial intelligences and A.I.-like systems would be easier to create with a memristor-based system~\cite{NS2009}. Memristor-based systems have other advantages over transistor-based systems: their memory is not volatile, and thus removal of the power source does not erase data~\cite{Chua1976}. From a green computing point of view memristors only draw power when accessed. Again, this is similar to the brain, and fits well with neuromorphic computing paradigms where we are concerned with the propagation of spiking direct current signals rather than repeated A.C. clock cycles. 

Thus far, we~\cite{David1,David2} and others~\cite{Neuro2011} have focused on utilising memristors in spiking networks, using a bottom up approach to building memristor computers. Such networks are often considered in terms of their use as control systems for autonomous or distributed systems and accordingly we focused on path-finding by a single autonomous agent. While the bottom-up approach is good for increasing complexity of a single agent, we now present the complimentary approach which has uses for a different class of problems. 

Memristors can be used at a higher level of abstraction such as in a top-down approach to model the environment. An example of this is maze-solving where the maze is modeled as a grid of memristors~\cite{PVMaze} (note path-finding via grids of resistors had been done prior to this~\cite{Marshall1991,Blake1991}.). The maze can be solved in the time taken for one memristor to switch, regardless of the length and solution path. This method has only been simulated, but the real interest lies in using the memristors in hardware to allow the laws of physics to solve the problem very quickly. 

Gathering as a process can be conceptually separated into two parts: firstly finding the resource (resource location and path finding) and secondly harvesting or gathering the resource (following laid-down paths and returning with resources). Here we concentrate here on the problem of efficient gathering in an environment in which the resources have been located. The gatherers under study could be ants gathering food or autonomous agents mining useful resources for a factory or collecting samples for scientific research.  

\subsection{Ants}

The existence of pheromones and their use in guiding ants was first reported in 1880~\cite{Lubbock1880} and positive reinforcement of ant trails was reported in 1962~\cite{Wilson1962}. Since then the picture has become much more detailed: there are several pheromones~\cite{ants8} which last for different times~\cite{ants9} with some that reverse previous instructions~\cite{ants7} that together tell the ants where to go and there are additional geometrical cues to tell them where they are~\cite{ants6}. Nonetheless, the simpler model of ants exploring their environment and responding by positively reinforcing pathways has given rise to several different ant optimisation algorithms from the original~\cite{DMandC1996}, to Ant Colony Optimisation for optimisation problems~\cite{DDC1999} and extensions of that to the Ant Colony System~\cite{ACS} and Rank based Ant system~\cite{Rank}. These algorithms have been successfully applied to various np-hard problems such as the traveling salesman~\cite{ants10}.

Given that study of ant searching behaviour has given rise to such a range of useful computational techniques, we chose to ask what they can teach us about harvesting. Different ant species have different behaviours, but in general they seek to maximise the colony's energy intake~\cite{ants2}, usually focusing on energy maximisation rather than harvesting time minimisation. For example, \textit{Solenopsis germinata} will preferentially go to the closer, bigger and higher concentration food sources over the more distant, smaller sources~\cite{ants2-100}. The Leafcutter ants, \textit{Atta}, have a slightly different technique, they preferentially go to only the best leaf site when they are in a resource rich environment and spread out to a greater diversity of trees/sites when in a resource poor environment\cite{ants2-86}. 

Some ants separate the finding of resources from their gathering. For example, once seed-gathering ants have found a food source, they will return to search that area, completely ignoring any seeds placed near their path~\cite{ants14}. Similarly, Leafcutter ants will create trails which the gatherer tend to stick to. Thus, it is valid to concentrate on the behaviour of gatherers once the environment has been explored. 

The dynamics of ant colonies are non-linear, which makes memristors, as non-linear devices, well suited to model them. To demonstrate the usefulness of memristors as an environmental model, we look at the problem of the most efficient gathering techniques in two very different situations, that of a resource rich and resource poor environments. We will focus on three different models of behaviour: 1, the Sequential Gathering technique where all the gatherers go to each food source in turn, starting with the best; 2, the All Sites model where the gatherers are split between all the food sources; and 3, the Leafcutter model where all the ants deplete the best resource and then spread out amongst the rest.

\section{The Memristor Model of the Environment}

\subsection{The Model Memristor}

Memristance $M$, relates charge, $q$ and magnetic flux $\varphi$ by $\varphi = M(q) q$. As $M$ is a function of $q$, the memristance is controlled by the amount of charge on the memristor, this is related to the current that has passed through it, and so $q$ the memory of the device. $q$ is a function of time, $t$, and at any instant in time the memristor acts like an Ohmic resistor in that $V = R I$, where $R$ is the instantaneous resistance, and memristance is the time-varying resistance.

Memristors have physical limits to the possible resistance values: the lower limit $\Ron$ and the upper limit $\Roff$. The first model of memristance to relate it to physical measureables~\cite{Nature2008} gave the memristance as

\begin{equation}
M(q) = \Roff - \Roff \Ron \beta q ,
\end{equation}

where $\beta$ is $\frac{\mu_v}{D^{2}}$. This model refers to the Strukov memristor~\cite{Nature2008} where two different forms of titanium dioxide (with differing resistivities) are interconverted based on the drift of oxygen vacancies, as governed by the oxygen vacancy ion mobility, $\mu_v$, across a thin-film of thickness $D$. However, for our current model, we can view $\beta$ as a parameter which varies based on the material properties of the device (for a discussion on the effect of varying this parameter, see~\cite{David2}). 

\subsection{Methodology}

\subsubsection{Using Memristors to Model the Environment}

The test environment is five different food sources different distances from the nest which is modeled by five memristors in series with a 5V potential difference applied across them. The total voltage drop does not change and relates to the number of available gatherers in the system. The memristors are set to the ON state (the low resistance mode) so that an increase in charge will eventually switch the device into the OFF state. Thus, for ease of use, the memristance is modeled as $M(q)=\Ron + \Roff \Ron \beta q$, which is equivalent to the equation above if we include the boundary condition: $\Ron \leq M(q) \leq \Roff$.

At the start of the simulation, each memristor is charged to a different degree, ie $M(t=0)$ varies, and this number encompasses the amount of resource, and the length and difficulty of the path (the more resources present and the easier the path, the lower the value of $M(t=0)$). This is done by changing $\Ron$ in the model, in the lab this would be done by charging the memristor up by a set amount before the start of the experiment. $\beta$ is set to 1 in this model for all memristors, and it represents the difficulty getting resources from the site. The memristor equation gives a curve, and $\beta$ controls the curvature~\cite{David2} and due to the non-linearity of the curve this includes the law of diminishing returns whereby, because gatherers get in each other's way, each additional gatherer at a resource gives a smaller productivity gain as the number of gatherers at a resource rises. $M$ increases as the food is removed, until it hits $\Roff$, at which point the resource is considered depleted (the amount of resources at the start can vary, but when depleted, it is depleted no matter how much there was to begin with). As $M$ has a hard top limit, the resources are scaled to start in different places. The sum of the current is the rate of resource influx at the nest. In this situation, a real memristor system would not have a limited current, but the total amount of resources in a test environment should be the same, thus we use

\begin{equation}
\mathrm{\% \, of \, gathered \, resource \, on \, step \,} n = \sum_{t=0}^{t=n} \frac{I(t)}{I(D)}
\label{eq:I}
\end{equation}

where $I(D)$ is the current on the step where the environment is entirely depleted. 

For the rich environment, the memristors are set to: $M_1(0)=1\Omega$, $M_2(0)=2\Omega$, $M_3(0)=0.5\Omega$, $M_4(0)=15\Omega$ and $M_5(0)=4\Omega$. The poor environment contains one very good site $M_1(0)=0.5\Omega$, and the rest are set to $60\Omega$, $70\Omega$, $80\Omega$ and $90\Omega$ respectively (the order does not matter). Although these are given in terms of $\Omega$ for convenience, because $\beta$ is a changeable parameter in this model these are really reduced units.


\subsubsection{Using Memristors to Model Different Gathering Techniques}

We simulated a series circuit of 5 memristors to model the ants going to all the sites, and this is the All Sites model. The conceptually simplest gathering technique, the Sequential model, is to send all the ants to each food site in decreasing order of their richness. To calculate this, each of the memristors was charged up individually until it reached $\Roff$, whereupon the next memristor was charged. To make it a valid comparison, the single memristors were left drawing current once they had reached $\Roff$ as this could happen in the other two models (the value of this current draw is small). The Leafcutter model is related to the All Sites model: the best memristor is run singly, then when depleted, it's left drawing a minimal current whilst the All sites model is applied to the remaining 4 memristors. 

\section{Results}

\subsection{Example of how the Memristance and Voltage Operates in the All Site Modeled Electronic System}

\begin{figure}
	\centering
		\includegraphics[width=12cm,height=7.25cm]{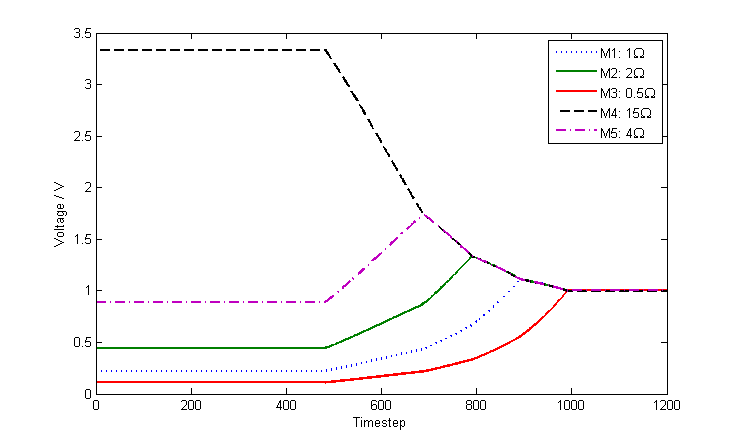}
	\caption{The voltage across each memristor as a function of time. These curves show how ants are spread between the food sites. All sites are depleted when the voltage drop across all the memristors is equal.}
	\label{fig:SeriesVoltage}
\end{figure}

This discussion uses a graph taken from the All Site model in the rich environment simulation, the same principles apply in the poor environment simulation. The memristance starts at $\Ron$ and changes as a function of the integral of the current passed through the system until it reaches $\Roff$. In terms of the gathering model, the lower $\Ron$, the more resources available at that position initially. As $\beta$ is the same for each memristor in this example, the curvature of each memristor is the same and hence the rate of increase in difficulty in resource gathering is the same across all the sites. For this reason, the resistance change does not effect the voltage under 500 steps (see figure~\ref{fig:SeriesVoltage}), because the voltage is shared between memristors in proportion to their relative resistances.


Figure~\ref{fig:SeriesVoltage} shows how the voltage drop across each memristor changes with time step. Once the 4th food source has been depleted, its share of the workers drops drastically, with most going to the next closest to being depleted, site 5, and when this is depleted, most workers move to the next resource. This continues with the number of workers assigned rising on the remaining resources as others are depleted, until time step 992 when all the resources are depleted. 

\section{Comparison Between Gathering Approaches}

\subsection{Resource Rich Environment}

\begin{figure}[htb]
	\centering
		\includegraphics[width=10cm,height=7.25cm]{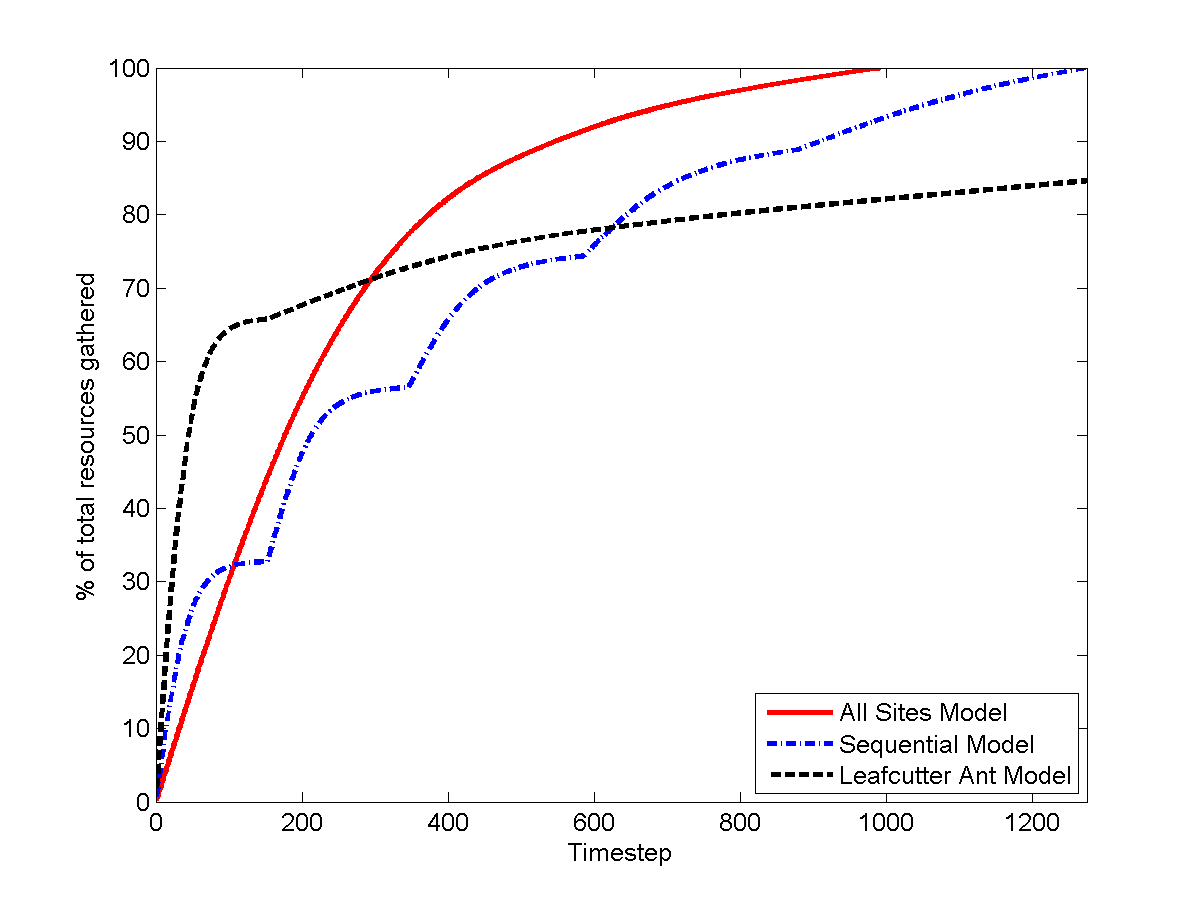}
	\caption{Comparison of three gathering models. The Leaf cutter model is better in the short terms but takes 2978 time steps to reach 100\% (all resource consumed), this is not shown in this figure.).}
	\label{fig:GatheringComparison}
\end{figure}

As the maximum value of $M$ is 100$\Omega$, even the highest memristor starting value, 15$\Omega$, still represents a good resource site. The rate of resource influx at the nest is the normalised cumulative current which is plotted in figure~\ref{fig:GatheringComparison}. 

The worst model, in our opinion, is the Sequential model: it takes longer than the All Sites model, and although it beats it for resource influx in the short-term this advantage is lost after a very short time. It depletes the total environment quicker than the Leafcutter allocation method, but does not have that model's advantage of a high rate of initial influx.

To answer which is the best gathering method, we first need to ask what we expect from the best method, the quickest influx of resources or the shortest time to completely deplete the surrounding environment. The Leaf Cutter model has a clear advantage at short-times because the gatherers are concentrating on the best resource in the system. This has a cost: the Leaf-cutter model takes much longer than the other two, 2978 time steps, compared to 967 for the All Sites model and 1274 for the Sequential model, which is 108\% longer than the shortest time. There is a finite amount of resources, but due to the diminishing returns, it can take different amounts of time to get them all and the measure of this difference is the differing efficiencies of the gathering techniques. In terms of fully depleting the environment, the Leaf Cutter allocation method is less efficient in this environment.

This fits perfectly with the entomological observation that ants aim to optimise the maximum energy/resource intake rather than minimize the time taken. Also, the Leafcutter method is based on the ant's strategy when in rich environments, so we might expect this model to fare well in this simulation. In the wild, there are advantages to gathering the best resource first, it prevents competitors from taking it and mitigates against any change in the environment. 

However, in a rich environment where the gatherers can expect to be relatively undisturbed by competitors the All Sites model is the better approach as it allows the gatherers to entirely deplete the environment in under half as much time. Although the Leafcutter model is quicker at the start (taking 44 steps to gather 50\%), the All Sites is still fast having gathered 50\% of the total resources in under a 5th (17\%, 173 timesteps) of the total time. Similarly, although the Sequential model is quicker than the All Sites at the start, the productivity gains trail off quickly (over 200 time steps). 

Both the Leafcutter model and the Sequential model start with the best resource site first. This may seem like an obviously good idea, and interestingly, it is not how the All Sites model works. Instead the majority of gatherers go to the worst resource site first! This is because the difficulty of gathering increases with time as the `low-hanging fruit' is taken. Thus, even the worst site yields its best resource output to begin with, and the distance to it is compensated by the high productivity at the other sites. 

Once the worst resource site is depleted, that frees up workers to go to the other sites and compensate for decreasing productivity at those. Thus, if the resources are in an environment are known, it is more efficient to spread agents out amongst all of them, (with more to the worse resources) than to deplete each resource in turn. Even depleting the best resource first will delay the time taken to gather from the rest due to diminishing returns. Note, if it is desirable to try to regulate resource influx the All Sites model would be the best choice. 

\subsection{Resource Poor Environment}

\begin{figure}[htb]
	\centering
		\includegraphics[width=10cm,height=7.25cm]{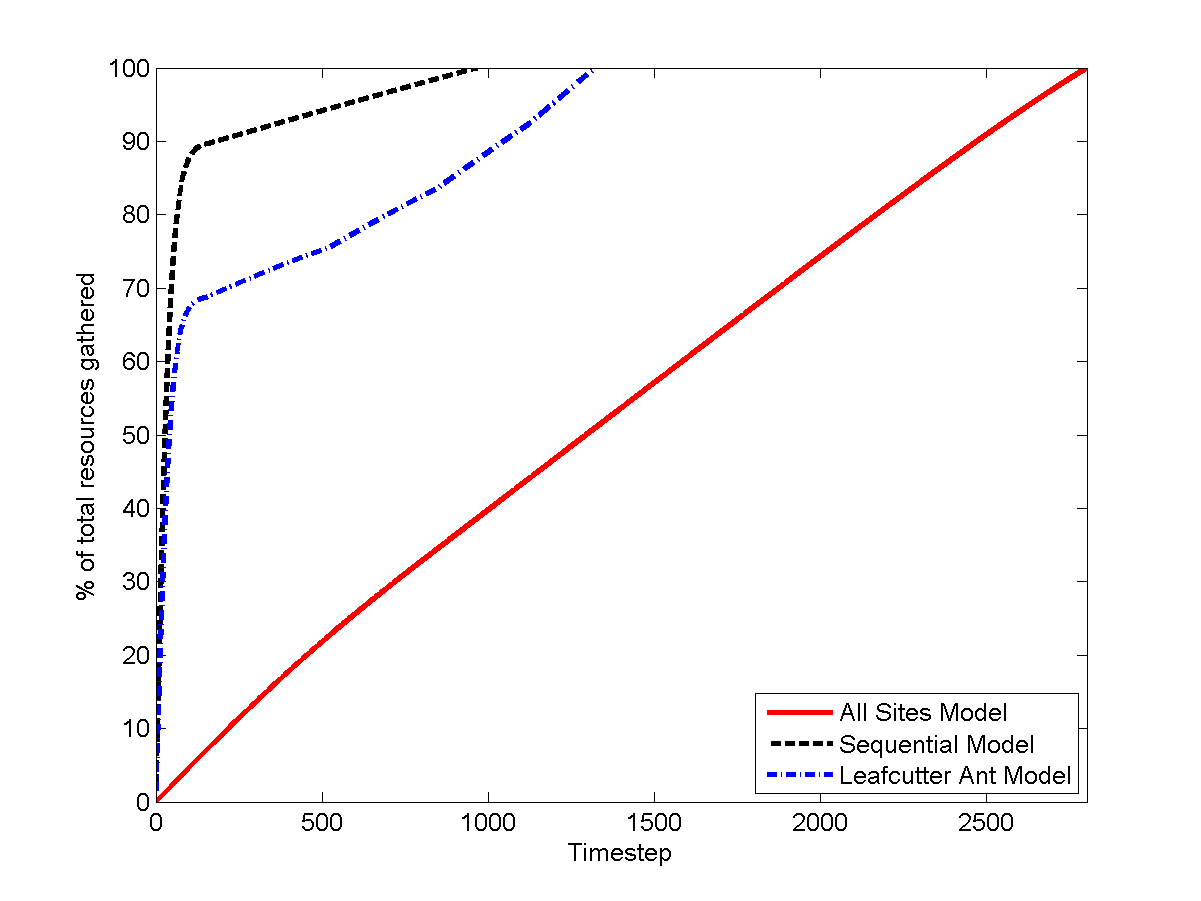}
	\caption{How different models fare in the resource poor environment. Despite being based on ant behaviour in a rich environment, the Leafcutter ant model fares best.}
	\label{fig:LeafAntMemComparison2}
\end{figure}

In this environment, the All Sites model is not very good, as it drastically slows the time taken to deplete the good resource and the environment (both to 2801 steps), see figure~\ref{fig:LeafAntMemComparison2}. However, the Leafcutter and the Sequential models both do much better by depleting the best source first, they get 90\% and 68\% in their initial surge ($\approx$ the first 100 timesteps). The Leafcutter model beats the Sequential model on the time taken to deplete the entire environment by 967 timesteps to 1321 and is overall more efficient. This is surprising as the Leafcutter model was based on the ant behaviour in a rich environment. Real Leafcutter ants would spread their focus amongst all the sites in a poor environment, this suggests that in such a situation ants are allocated differently to All Site allocation model. 

The Leafcutter allocation model is clearly better than the Sequential model despite utilising the All Sites model over the latter time steps. This suggests that when the food sources are all of a similar magnitude (the rich environment was arguably close to this), it is better to send gatherers to all of them, weighted to the smallest, easiest depleted sources. If the sites are similar to each other this results in productivity gains whereby the slowing of productivity at one site is compensated by the increase at another. When the resource sites are vastly different, equalising the resource influx limits the best resource to the speed of gathering from the worst.

\section{Conclusions}

There are a few competing considerations when deciding how to spread gatherers around. More gatherers will get more resources from a single site, and the fewer resource sites there are the more gatherers can be put on each one. Thus, depleting the smaller sites first will give more gatherers to the pool to be reassigned to another resource site, increasing efficiency when the resources are similar in size. However, the extra amount of work added per gatherer decreases with the number of gatherers, so it makes sense to spread the gatherers around the sites. Finally, the rate of output at each site changes as that site is worked due to an increase in difficulty gathering and gatherer-gatherer interactions. Knowing how best to allocate gatherers in a dynamically changing system is difficult: the memristor based All Sites model can help balance out productivity and speed up gathering when resources are similar and the ant and memristor based Leafcutter model can be helpful in a mixed environment. Thus, these algorithms may have some use in both work allocation and the study of social insects. ering when resources are similar and the ant and memristor based Leafcutter model can be helpful in a mixed environment. Thus, these algorithms may have some use in both work allocation and the study of social insects. 

Memristors can be synthesized and thus all of these calculations can be done quickly in hardware and this is the main interest in using this technique. In this paper a complex environment has been modeled with only one memristor per path, making this a very lightweight model, so there is room for combinations of memristors which could allow the modelling of more realistic scenarios.

The similarity between memristors and biological components has been well observed. There may also be similarities between memristors and higher levels of biological organisation. It is known that many ants can produce more complex behaviour than study of a single ant might lead us to believe and study of this phenomena has been extended to our own species~\cite{ants4}. Ant colonies have short-term collective memories, their complex patterns of pheromones act as stored information, so it is possible that memristors could be used to model this institutional memory. A new model for the ant's behaviour that focuses on institutional memory might shed light on the more complex institutional memory which differentiates human societies and separates us from hunter-gatherer societies.

\subsubsection*{Acknowledgments.} E.G. would like to acknowledge support from EPSRC grant EP/H014381/1.
%
%

%
\end{document}